\documentclass{Interspeech}

\usepackage{array}

\usepackage{comment}

\usepackage{tabularx}
\usepackage{adjustbox}

\usepackage{helvet}

\usepackage{hyperref}

\usepackage{graphicx}

\usepackage{amsmath}

\usepackage{times}
\usepackage{latexsym}

\usepackage[T1]{fontenc}
\usepackage[utf8]{inputenc}

\usepackage{microtype}

\usepackage{xcolor}
\usepackage{url}
\usepackage{booktabs}

\newcommand{\natalie}[1]{\textcolor{magenta}{Natalie: #1}}

\interspeechcameraready

\title{Pitch Accent Detection improves Pretrained Automatic Speech Recognition}

\author[affiliation={1}]{David}{Sasu}
\author[affiliation={2}]{Natalie}{Schluter}

\affiliation{Computer Science}{IT University of Copenhagen}{Denmark}
\affiliation{}{Apple}{Denmark}
\email{dasa@itu.dk, natschluter@apple.com}

\usepackage{comment}

\begin{document}

\maketitle

\begin{abstract}
    
We show the performance of Automatic Speech Recognition (ASR) systems that use semi-supervised speech representations can be boosted by a complimentary pitch accent detection module, by introducing a joint ASR and pitch accent detection model. The pitch accent detection component of our model achieves a significant improvement on the state-of-the-art for the task, closing the gap in F1-score by 41\%.  Additionally, the ASR performance in joint training decreases WER by 28.3\% on LibriSpeech, under limited resource fine-tuning.  With these results, we show the importance of extending pretrained speech models to retain or re-learn important prosodic cues such as pitch accent. 
\end{abstract}

\section{Introduction}
Models based on self-supervised speech representations have in recent years claimed state-of-the-art performance in ASR \cite{baevski2020wav2vec,hsu2021hubert}.  Moreover, they have been permitted to bypass both a heavy speech-science informed featurisation component, as well a language dependent acoustic dictionary resource writing component.  In doing so, such models have become seemingly less human reliant during development, provided adequate quantities of raw speech data and computational resources.  

However, the training techniques for these self-supervised speech models do not reveal what exactly is deemed important by these models and later retained within their output speech (SSL) representations.  Subsequent studies have since introduced benchmarks and metrics to analyse the linguistic knowledge of these models at different levels, mostly from the point of view of assessing the existence of this linguistic knowledge. For example,  the Zero-Resource Speech challenges\footnote{\url{https://zerospeech.com}} provide tests beds to analyse phonetic, lexical, syntactic, and semantic level book-keeping of these representations \cite{zerorec2017,zerorec2020,zerorec2021}.  More recently, ProsAudit was introduced to provide a similar book-keeping of the prosodic information retained in these speech representations \cite{deseyssel23_interspeech}.  However, these studies and corresponding benchmarks do not elucidate whether refocusing these SSL representations to retain more of the original linguistic signal could correspond to better performance downstream in basic but central speech tasks like ASR.

One main bottleneck to carrying out such a study is the sheer scarcity of datasets with prosodic annotations, and the previous expense in generating these annotations using trained linguists.  One can, however, envisage a scheme for obtaining simple prosodic annotations for ``important'' words--pitch accent annotations--in an utterance from non-specialists, and even for non-written languages, by which annotators simply press a button when emphasised segments of an utterance are heard.  Still, currently, the only existing datasets for English (and any other language) are relatively small--the largest being the Boston University Radio News Corpus for English with 11 hours of data \cite{ostendorf1995boston}.  Is there any benefit for spoken language understanding tasks in extending current and/or developing new prosody datasets?  To assess this question, one must investigate the role of prosody in these tasks.  
%\footnote{* Currently at Apple.}

\subsection{Contributions} In this paper we study the role of prosodic information, specifically focusing on pitch accents, in ASR.  Our contributions are as follows.
\begin{enumerate}\setlength{\itemsep}{-0.5pt}
    \item We streamline and significantly boost the performance of the current state-of-the-art model for pitch accent detection.
    \item We present a multi-task model for integrating pitch accent detection into the ASR task, which improves the performance for ASR in limited resource settings. %both tasks (over BURNC....).
    \item We then automatically annotate the pitch accents of a small dataset using self-training, and then apply it in our proposed joint model, achieving even further ASR performance boosts.\footnote{We make all our code freely available at \url{https://github.com/sasudavid/prosodic_prominence_asr}.}
\end{enumerate}

\section{Related work}
%\noindent\textbf{Prosody Detection.}  
\subsection{Prosody Detection}  
There is a long line of research on automatic prosody detection (for example, \cite{taylor1995using, rosenberg2015modeling, shahin2016automatic,li2018automatic,stehwien2020acoustic,sabu2021cnn}).  With the advent of pretrained speech models, and in particular, wav2vec \cite{SchneiderBCA19} and wav2vec2 \cite{baevski2020wav2vec}, a new line of systems that builds on self-supervised speech representations has achieved the state-of-the-art in detecting prosodic boundaries in Czech broadcast news recordings \cite{kunevsova2022detection} and in pitch events and intonation phrase boundaries in English broadcast news \cite{Zhai_Wasegawa-Johnson2023}.  This latter model, called wav2TOBI, forms the point of departure for our multitask system presented here.  The question left open by these models, and others is whether these self-supervised representations adequately account for prosody, which has been shown to aid in an array of linguistic tasks like SLU \cite{noth2002use,shriberg2004prosody,shriberg1998can,rajaa2023improving,wei2022neural}, and parsing \cite{tran2017parsing,gregory2004sentence,kahn2005effective,dreyer2007exploiting,kahn2012joint,price1991use,beckman1996parsing}.  Or whether a specialised module should intervene and boost the prosodic signal for better performance.

%\noindent\textbf{Prosody with ASR.}  
\subsection{Prosody with ASR}  
In this paper, we are particularly interested in whether refocusing speech pretrained models on prosody might aid in performance for ASR.  Prosody has been previously shown to be of importance to ASR, both as engineered features, as well as through learning from prosody annotated datasets \cite{silverman1992towards, ostendorfy2003prosody,hirose2004use,hirschberg2004prosodic,hasegawa2005simultaneous,ananthakrishnan2007improved,vicsi2010using,chen2012new,kathania2020data,hasija2022prosodic,coto2021explicit}.  However, to our knowledge, there is no research that builds on pretrained speech models, whose application to pitch accent detection and ASR has resulted in the state-of-the-art performance.

\subsection{State-of-the-art ASR}
%\noindent\textbf{State-of-the-art ASR.}
%A summary of state-of-the-art performance in ASR over the Librispeech dataset is given at the end of this paper (Table \hyperref[Table_1]{~\ref*{Table_1}}) for comparison with our work presented here.  
Previous work such as \cite{baevski2019effectiveness, baevski2020wav2vec, hsu2021hubert, ling2020decoar,likhomanenko2020slimipl, xu2020iterative, park2020improved} demonstrate state-of-the-art performance in ASR over the Librispeech dataset.
For model comparability in the work presented here, we focus on the wav2vec2 model, which is the pretrained model fine-tuned by the wav2TOBI model for prosody detection.%, and which is the model on which we base our system presented here.

\section{Modelling prosody and ASR}
\subsection{Datasets}
Our research uses the Boston University Radio News Corpus (BURNC) \cite{ostendorf1995boston}, Librispeech \cite{panayotov2015librispeech} and Libri-light \cite{kahn2020libri} corpora.

The BURNC dataset is a broadcast news-style read speech corpus which contains 11 hours of speech, sourced from 7 different speakers (3 female and 4 male).
It consists of audio snippets with their transcriptions, phonetic alignments, parts-of-speech tags and prosodic labels. We used 75\% of the data in this dataset for training, 15\% for development and 10\% for testing.  Because multiple readers may have read the same news story in BURNC, we ensure that no news stories appearing in the test set also occur in the training set.

We represent pitch accent labels from BURNC following the binary labelling strategy presented in \cite{Zhai_Wasegawa-Johnson2023}. We assign positive labels to time-frames corresponding to audio segments labelled in BURNC as having pitch accents, and negative labels otherwise.
Following \cite{Zhai_Wasegawa-Johnson2023}, we preprocess BURNC audios by splitting them into overlapping clips of 20s, at 10s intervals.
The Librispeech dataset consists of 1000 hours of audio samples sourced from the LibriVox Project.
In our work, the dev-clean and test-clean data subsets were used for model development and evaluation, respectively. 

The Libri-light dataset is made up of 60,000 hours of audio and, similarly to the Librispeech corpus, was also sourced from the LibriVox Project. We used the Libri-light limited resource training data subsets, namely, train-1h (LS1), which consists of 1 hour of labelled audio data. 

\subsection{A joint model for prosody and ASR}
\begin{figure}[h]
    \caption{Joint Prosody-ASR Model}
    \label{fig:1}
    \centering
    \includegraphics[width=0.4\textwidth]{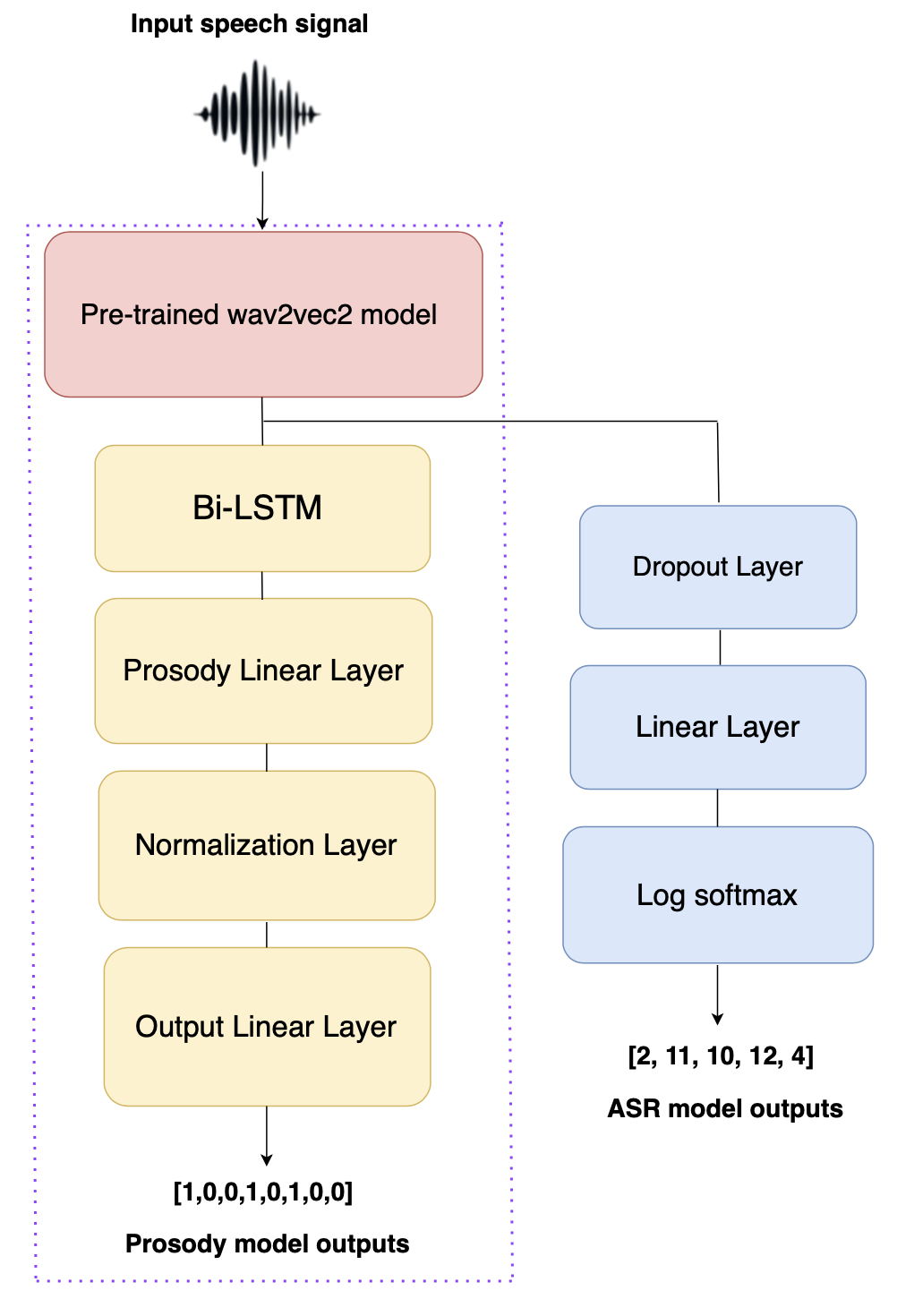}
\end{figure}

Our proposed system jointly learns pitch accent detection and automatic speech recognition (cf \hyperref[fig:1]{Figure~\ref*{fig:1}}), using the pitch accent annotations from BURNC.  

For both ASR and prosody detection, raw audio input is sent through the pretrained wav2vec2 model \cite{baevski2020wav2vec} with a language modelling head on top for Connectionist Temporal Classification for the ASR task.
For pitch accent detection, we built upon the prosodic event detection model proposed by \cite{Zhai_Wasegawa-Johnson2023}, wav2TOBI.  In wav2TOBI, wav2vec2 timestep representations are concatenated with fundamental frequency features and fed through a BiLSTM, followed by a classification layer, with mean-squared loss.
Our model streamlines wav2TOBI in the sense that we no longer require fundamental frequency features and make pure use of wav2vec2 output representations. On the other hand, we introduce an extra linear layer followed by layer normalisation before the classification layer.

We train our proposed joint model by minimising a joint loss function $\mathcal{L}_j= \mathcal{L}_{asr} + \mathcal{L}_{pad}$, which is the combination of the ASR model loss $\mathcal{L}_{asr}$ and pitch accent detection model loss $\mathcal{L}_{pad}$.

\noindent\textbf{Results for prosody detection.} In the single task setting, for prosody detection, these simple changes result in significant improvements in pitch accent detection performance over wav2TOBI, even without recourse to the additional fundamental frequency features (Table \hyperref[Table:prosody_detect_perf]{~\ref*{Table:prosody_detect_perf}}).

\begin{table}[ht]
\caption{Prosody detection system performance at varying levels of error tolerance (Tol) in milliseconds.  Our basic model (Ours Prosody) outperforms the state of the art wav2TOBI system in pitch accent detection.  Our semi-supervised training approach (Ours Semi-Sup) further improves performance.}
\label{Table:prosody_detect_perf}
\centering
%\begin{adjustbox}{width=0.7\textwidth}
    \begin{tabular}{l| c c c c}
        \toprule
        \textbf{Model}&\textbf{Tol}&\textbf{Prec}&\textbf{Rec}&\textbf{F1}\\
        \midrule
        wav2TOBI& 0 ms & 0.13 & 0.11 & 0.12\\
        & 40 ms & 0.70 & 0.61 & 0.65\\
        & 80 ms & 0.87 & 0.74 & 0.79\\
        & 100 ms & 0.89 & 0.76 & 0.81\\
        \hline
        Ours Prosody& 0 ms & 0.37 & 0.36 & 0.36\\
        & 40 ms & 0.82 & 0.8 & 0.81\\
        & 80 ms & 0.89 & 0.86 & 0.87\\
        & 100 ms & 0.9 & 0.87 & 0.88\\\hline
        Ours Semi-Sup & 0 ms & 0.49 & 0.48 & 0.48\\
        & 40 ms & 0.85 & 0.82 & 0.83\\
        & 80 ms & 0.92 & 0.88 & 0.9\\
        & 100 ms & \textbf{0.93} & \textbf{0.89} & \textbf{0.9}\\

        \bottomrule
    \end{tabular}
%\end{adjustbox}
\end{table}

\begin{comment}
    
\begin{table}[ht]
\centering
\begin{adjustbox}{width=0.25\textwidth}
    \begin{tabular}{c c c c}
        \hline
        \textbf{Tol}&\textbf{Prec}&\textbf{Rec}&\textbf{F1}\\
        \hline
        0 ms & 0.49 & 0.48 & 0.48\\
         40 ms & 0.85 & 0.82 & 0.83\\
         80 ms & 0.92 & 0.88 & 0.9\\
         100 ms & 0.93 & 0.89 & 0.9\\
        \hline
    \end{tabular}
\end{adjustbox}
\caption{Performance of our semi-supervised prosody annotation approach}
\label{Table_3}
\end{table}
\end{comment}

\subsection{Semi-supervised Prosodic Event Detection}
Our joint prosody-ASR modelling is limited to datasets where these prosodic labels are available.  In order to address ASR performance for a dataset like LibriSpeech, where prosodic labels are unavailable, we resort to semi-supervision--specifically, self-training with model voting.  In these experiments, we focus on the smaller Libri-light train-1h (LS1) dataset in order to minimise the possible extrapolation error of a larger dataset.

%Before we train our joint Prosody-ASR model, we utilise a semi-supervised technique to provide pitch accent labels for our audio data. To obtain the best possible pitch accent labels for our audio data, we experimented with different semi-supervised labelling techniques and used the labels provided by the best performing technique to train the Joint Prosody-ASR model. We selected BURNC train dataset as well as the Libri-light train-1h to develop our semi-supervised labelling approach and used BURNC test dataset to test its efficacy. 

%In our chosen semi-supervised labelling technique, we first 
We partitioned BURNC train set into three subsets as possible hold-outs.  For each hold-out subset, we used the remaining 2/3s of the original train set to retrain a new model.  We used each of the three models to obtain three predicted labels for each instance of the LS1 train set, and retained the majority class label of these for each instance.  The prosody labelled version of LS1 was then added to the full BURNC train set, and then checked over BURNC test set for performance gains.  If there were gains, we repeated the process now with the prosody labelled LS1 as part of the partitioning step, replacing the labels of LS1 at each iteration.  Otherwise the process halts. Our training process halted after 4 iterations.\footnote{Note that we tried a number of different self-training techniques, but this simple voting technique worked the best.}

\noindent\textbf{Results for prosody detection.} The single task results for this approach on prosody detection are given in Table \ref{Table:prosody_detect_perf}.  We observe that across all measures and error tolerances, this method improves performance and achieves, to our knowledge, the current state-of-the-art.

%\{\emph{$S_1$},\emph{$S_2$},\emph{$S_3$}\} of equal size and then combine these different subsets into different dataset combinations, consisting of 2 out of 3 of these different subsets. This generates 3 different dataset combinations, \{\{\emph{$S_1$}, \emph{$S_2$}\}, \{\emph{$S_1$}, \emph{$S_3$}\}, \{\emph{$S_2$}, \emph{$S_3$}\}\}, which we then used to train 3 different prosody labelling models, \emph{$P_1$}, \emph{$P_2$} and \emph{$P_3$}. We then apply these models to generate pitch accent labels for the Libri-light train-1h dataset and select the majority class of these labels as the final predicted labels for each audio in the Libri-light train-1h dataset. The prosodically labelled Libri-light train-1h dataset is added to the BURNC train dataset and used to train another prosody labelling model \emph{$P_*$} which is then evaluated on the BURNC test dataset. This entire process is repeated until there is no further improvement in the performance of the prosody labelling model \emph{$P_*$}. Its performance when evaluated is showcased in Table \hyperref[Table_3]{~\ref*{Table_3}}.

\section{Experimental setup and results}

We use the base-960h wav2vec2 pretrained model\footnote{\url{https://huggingface.co/facebook/wav2vec2-base-960h}}.  Our models are all trained for 30,000 steps on three v100 GPUs, using default parameters, and take approximately 8 hours to train.\footnote{Default parameters are from \url{https://huggingface.co/docs/transformers/en/model_doc/wav2vec2\#transformers.Wav2Vec2ForCTC}.}  Results for word and character error rates (respectively WER and CER) are given in Table \ref{Table_4}.  All models were fine-tuned for ASR (resp. ASR and prosody jointly) on LS1.  Following \cite{baevski2020wav2vec}, and for comparability with state-of-the-art ASR models, we also experiment with 4-gram\footnote{\url{https://github.com/kpu/kenlm}} and Transformer \footnote{\url{https://github.com/facebookresearch/fairseq/blob/main/examples/language_model/README.md}} language models.

In the fine-tuning process, following \cite{baevski2020wav2vec}, the pretrained model remains frozen during the first 15K steps, after which the entire model is trained for the remaining 15K steps. The feature encoder remains frozen throughout fine-tuning. For the Prosody-ASR model, we use the prosody labelled version of LS1 outlined above. 

\begin{table}[th!]
    \caption{WER and CER results for ASR all with the wav2vec 2.0 speech model.}
    \label{Table_4}
    \centering
%\begin{adjustbox}{width=0.5\textwidth}
\begin{tabular}{p{2.5cm} | cc|cc}
        \toprule
        &\multicolumn{2}{c|}{LibriSpeech}&\multicolumn{2}{c}{BURNC}\\        
        \textbf{Model}&\textbf{WER}&\textbf{CER}&\textbf{WER}&\textbf{CER}\\
        \midrule
        \multicolumn{5}{c}{ASR-only training}\\\hline
        no LM & 6.0 & 1.0 & 23.0 & 7.0 \\
        4-gram LM & 5.0 & 1.30 & 17.59 & 6.15 \\
        Transformer LM & 5.0 & 1.50 & 18.0 & 5.0 \\
        \hline
        \multicolumn{5}{c}{joint ASR-prosody training}\\\hline
        no LM & 4.3 & 1.1 &  20.0 & 7.7\\
        4-gram LM   & 3.0 & 1.0 &  13.25  & 4.0\\
        Transformer LM   & 4.0 & 1.1 & 14.9  & 4.0 \\
        \bottomrule
    \end{tabular}
%\end{adjustbox}
\end{table}

\noindent\textbf{Results.} 
Our findings indicate that while character error rate (CER) remains relatively stable, the incorporation of prosodic information in ASR training significantly enhances word error rate (WER) performance. Specifically, our joint ASR-prosody model achieves a 28.3\% reduction in WER on the LibriSpeech test set and a 13\% reduction on the BURNC test set when no language model (LM) is applied.
When utilizing a 4-gram LM, the joint ASR-prosody model further improves performance, yielding a 40\% reduction in WER on LibriSpeech and a 24.47\% reduction on BURNC, compared to the ASR-only baseline. Similarly, with a Transformer LM, we observe a 20\% WER reduction on LibriSpeech and a 17.22\% reduction on BURNC.
These results demonstrate that explicitly modeling prosodic features within the ASR training framework enhances the effectiveness of wav2vec 2.0 representations. Notably, the performance gains are consistent across different language modeling strategies, reinforcing the impact of prosody-aware training in improving ASR accuracy, particularly on the LibriSpeech benchmark.

\section{Error analysis and discussion}
We have shown above that pitch accent detection is useful for improving the performance of pretrained speech models in ASR tasks within limited resource scenarios. However, even though we improve upon the WER in most of the experiments that we perform with our proposed joint model, we notice that experiments that involve the BURNC dataset tend to on average have higher CER scores. We list two reasons for this phenomenon below and discuss their impact.

\noindent\textbf{Pre-processing mismatches and audio truncation.}
During the pre-processing of the transcriptions for the BURNC dataset, we transform the text into their uppercase representations and remove all punctuation marks that are not consequential in determining word meaning. For instance, given a word "CHIEF'S", we do not remove the apostrophe ( ' ) during pre-processing to form the word "CHIEFS" since doing so changes the inherent meaning of the word. Another example is "S.J.C's", we do not represent it as "SJCS". Even though this text pre-processing approach is well-warranted, it leads to higher CER scores during ASR since the BURNC dataset is filled with a plethora of acronyms, hyphenated and contracted words. Additionally, following the data pre-processing approach utilised in \cite{Zhai_Wasegawa-Johnson2023}, we split our audios into overlapping clips of 20s, at intervals of 10s, for input to the wav2vec2 model. This however leads to the truncation of words in the initial or final position. As a result of this, some of the words that are predicted by the model are incomplete and this leads to a higher CER score.

% We demonstrate an example of an output of our model where a predicted transcript contains the described problem:
%\textbf{correct text}:  WOOD CREDITS THE CITY'S \textbf{ANTI-CRACK} CAMPAIGN WITH KEEPING THAT DRUG UNDER CONTROL BOSTON HAS A CRACK PROBLEM 
%\textbf{predicted text}:  WOOD CREDITS THE CITY'S \textbf{ANTI CRACK} CAMPAIGN WITH KEEPING THAT DRUG UNDER CONTROL BOSTON HAS A CRACK PROBLEM

%\textbf{Audio Truncation}.
 %An example of an output of our model showcasing an instance of this problem can be seen below:

%\textbf{correct text}:  \textbf{MOST} CONTROVERSIAL CLAUSE IN THE SAFE 

%\textbf{predicted text}:  \textbf{ST} CONTROVERSIAL CLAUSE IN THE SAFE

\section{Conclusion}
In this paper we have presented an approach for leveraging prosodic information to improve the performance of a pretrained speech model in a limited resource scenario. 
The results from our experiments demonstrate that re-focusing self-supervised speech models on supra-segmental speech cues such as prosody could lead to significant performance gains in downstream tasks. 

%Since approximately 70\% of the world's languages are tonal in nature, with a wealth of usually untapped information residing in prosodic features, our work opens up the future for exciting research directions into the possible different ways in which such information could be deliberately exploited to improve speech systems.

\section{Limitations}
All experiments were carried out under the limited resource setting, with little fine-tuning data, due to the requirement of our method to use prosodic labels.  More work is required to investigate the real impact when fine-tuning with larger ASR datasets.  

Also, for prosodic cues, we only used pitch-accent, and with hard labels (0 or 1).  It is not clear whether other aspects of prosody would also be important for ASR.  This question remains open.
%\section{Acknowledgements}
%We would like to thank ITU High-Performance Computing Cluster for the computing resources.

\bibliography{thebib, natalieextras}
\bibliographystyle{IEEEtran}

\end{document}